\documentclass[letterpaper, 10 pt, conference]{ieeeconf}  

\IEEEoverridecommandlockouts                              

\usepackage{graphics} 
\usepackage{epsfig} 
\usepackage{mathptmx} 
\usepackage{times} 
\usepackage{amsmath} 
\usepackage{amssymb}  
\usepackage{makecell}
\usepackage{multirow}
\usepackage{booktabs}
\usepackage{colortbl}
\usepackage[table]{xcolor}
\usepackage[pagebackref,breaklinks,colorlinks]{hyperref}
\usepackage[capitalize]{cleveref}       
\usepackage{caption}
\usepackage{amssymb}
\usepackage{pifont}
\usepackage{stfloats}
\usepackage[caption=false]{subfig}
\usepackage{graphicx}

\newcommand{\cmark}{\ding{51}}%
\newcommand{\xmark}{\ding{55}}%

\newcommand{\myparagraph}[1]{\vspace{4pt}\noindent{\bf #1}}

\title{\LARGE \bf
Finetuning Pre-trained Model with Limited Data\\ for LiDAR-based 3D Object Detection by Bridging Domain Gaps
}

\author{Jiyun Jang, Mincheol Chang, Jongwon Park, and Jinkyu Kim
\thanks{J. Jang, M. Chang, and J. Kim are with Department of Computer Science and Engineering, Korea University, Seoul 02841, Korea.
}%
\thanks{J. Park is with Autonomous Driving Center, Hyundai Motor Company R\&D Division.}
}

\begin{document}
\definecolor{LightBlue}{rgb}{0.8235,0.9737,0.9882}
\definecolor{LightGrey}{rgb}{0.95,0.95,0.95}
\definecolor{White}{rgb}{1,1,1}
\newcommand{\red}[1]{\textcolor{red}{#1}}
\newcommand{\blue}[1]{\textcolor{blue}{#1}}
\newcommand{\green}[1]{\textcolor{green}{#1}}
\newcommand{\jinkyu}[1]{{\textcolor{blue}{[#1]}}}

\maketitle
\thispagestyle{empty}
\pagestyle{empty}


\begin{abstract}

LiDAR-based 3D object detectors have been largely utilized in various applications, including autonomous vehicles or mobile robots. However, LiDAR-based detectors often fail to adapt well to target domains with different sensor configurations (e.g., types of sensors, spatial resolution, or FOVs) and location shifts. Collecting and annotating datasets in a new setup is commonly required to reduce such gaps, but it is often expensive and time-consuming. Recent studies suggest that pre-trained backbones can be learned in a self-supervised manner with large-scale unlabeled LiDAR frames. However, despite their expressive representations, they remain challenging to generalize well without substantial amounts of data from the target domain. Thus, we propose a novel method, called Domain Adaptive Distill-Tuning (DADT), to adapt a pre-trained model with limited target data ($\approx$100 LiDAR frames), retaining its representation power and preventing it from overfitting. Specifically, we use regularizers to align object-level and context-level representations between the pre-trained and finetuned models in a teacher-student architecture. Our experiments with driving benchmarks, i.e., Waymo Open dataset and KITTI, confirm that our method effectively finetunes a pre-trained model, achieving significant gains in accuracy.


\end{abstract} 
\section{INTRODUCTION}
\label{sec:intro}

LiDAR-based 3D object detection has emerged as a fundamental task in autonomous driving (AD) and robotics, and recent works~\cite{second, pointpillars, center, pvrcnn++, pointrcnn, voxelnet} have achieved promising results. However, such models must be trained with large-scale annotated data, which is expensive and time-consuming. Moreover, their performance is often limited to in-domain data distribution, as discussed in literature~\cite{SN,st3d, 3dcoco, srdan,lidar-distill, dts} -- they may not adapt well to target domains with different sensor configurations (e.g., types of sensors, sensor's spatial resolution, density, and FOVs) or geometric location shifts (e.g., inferencing in different cities or countries). A common practice to address this issue would be new (large-scale) data collection and annotation in target domains, which are laborious and costly. Thus, it is highly demanded that a model can be continuously adapted well to target domains without needing large-scale annotated data.

\begin{figure}[t]
    \centering
    \includegraphics[width=.9\linewidth]{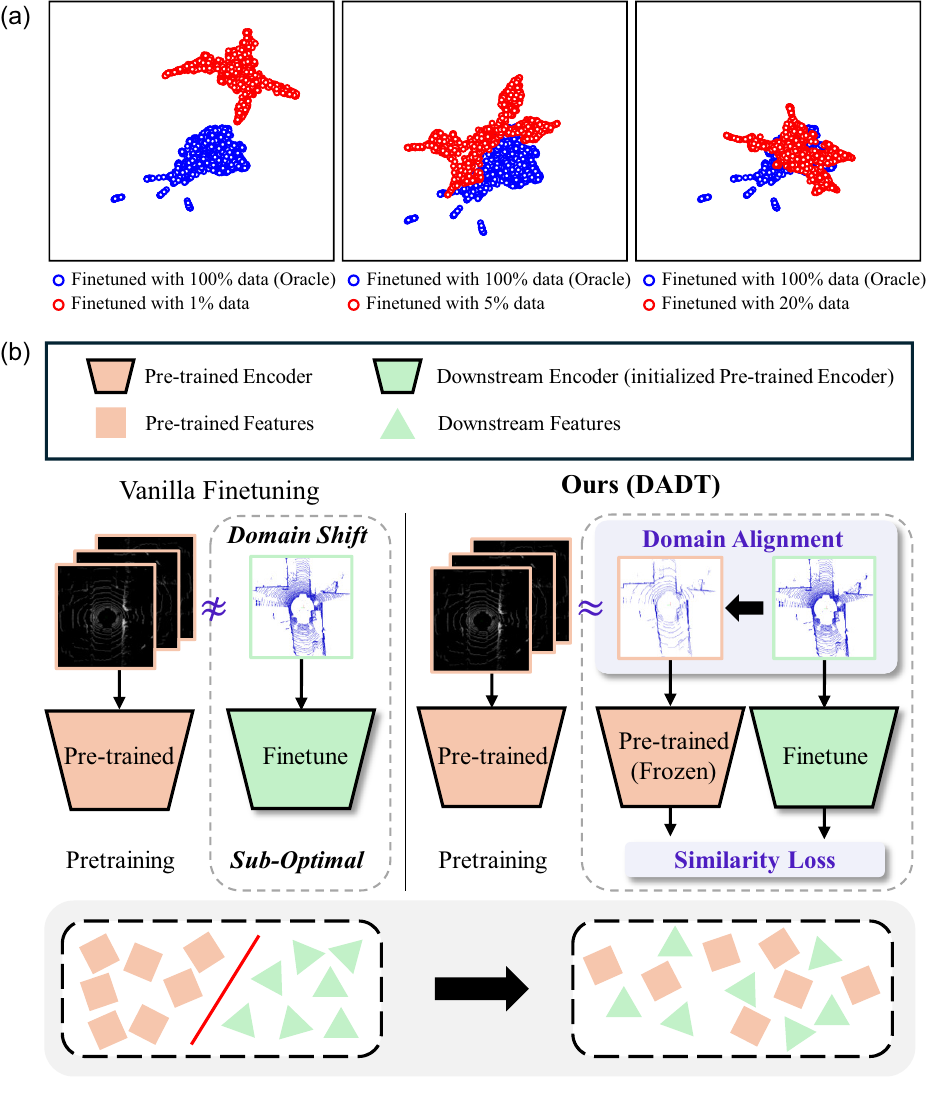}
    \caption{\textbf{(a) UMAP~\cite{umap} Visualizations.}  We compare 2D BEV features between (i) the oracle model (a pre-trained model by AD-PT~\cite{AD-PT} finetuned with the whole KITTI~\cite{kitti} training data) and (ii) similarly finetuned models but with smaller datasets. \textbf{(b) An  Overview of Architectures.} Conventional finetuning approaches (left) and our proposed Domain Adaptive Distill-Tuning (DADT) approach (right).}
    \label{fig:tsne}
    \vspace{-1.5em}
\end{figure}

Recent studies explored self-supervised representation learning on large-scale unlabeled point clouds~\cite{gcc3d, proposalcontrast, gdmae, voxelmae, AD-PT}, showing promising results in the 3D detection downstream task. For example, AD-PT~\cite{AD-PT} suggested a general representation model pre-trained with 1M unlabeled driving scenes from ONCE~\cite{ONCE}, followed by diverse data augmentation for robustness, showing notable performance on various AD benchmarks. However, we empirically observe that they still need substantial amounts of data to finetune the pre-trained model in a target domain (with domain gaps from source domains). For example, we provide UMAP~\cite{umap} visualizations in \cref{fig:tsne} (a) where we compare (i) 2D Bird's Eye View (BEV) features of AD-PT model finetuned with 100\% labeled KITTI~\cite{kitti} data (blue dots) and (ii) those from AD-PT models finetuned with different percentages (1\%, 5\%, 20\%) of labeled data (red dots). This demonstrates that the mismatch between pre-training (source) and downstream (target) datasets can negatively affect the performance of downstream tasks with limited data, sub-optimally leveraging the expressive pre-trained models.

We argue that a naive finetuning approach of a pre-trained model may harm the robustness of a model against distribution shifts, providing degraded performance (though the pre-trained model was trained with large-scale data). Inspired by recent work in the domain generalization task, we opt for a strategy where models learn similar features to those of approximation of ``oracle'' representations, which can be generalized well across any domain. Specifically, given a large pre-trained model as an approximation, we finetune a model with an objective of two components: (i) the original object detection task (i.e., Empirical Risk Minimization objective) and (ii) a regularization term between the pre-trained model (i.e., approximation of ``oracle'') and the target model. We simultaneously leverage the pre-trained model as the initialization and approximation of the oracle model. 

Here, as shown in \cref{fig:tsne} (b), we propose a novel finetuning approach called Domain Adaptive Distill-tuning (DADT), which is based on teacher-student architecture where a teacher network utilizes the frozen pre-trained backbone with density-aligned LiDAR inputs (target domain's LiDAR points are resampled to match those of source domain) and a student network finetunes its backbone (initialized from the pre-trained backbone) with the original density-non-aligned LiDAR inputs. We further use two BEV-based regularization terms, i.e., (i) object similarity loss and (ii) context similarity loss, to tie representations both for the teacher and student network together during the finetuning step, retaining oracle model's generalizable representations and preventing it from overfitting. Our extensive experiments with driving benchmarks, such as the Waymo Open dataset~\cite{waymo} and KITTI~\cite{kitti}, demonstrate that our method effectively finetunes a pre-trained model with limited target data ($\approx$ 100 LiDAR frames), achieving significant gains in accuracy. Our contributions are summarized as: 
%
\begin{itemize} 
    \item We propose a novel approach, called Domain Adaptive Distill-tuning (DADT), which aims to leverage and retain the representation power of a pre-trained model to effectively adapt to target domains with limited data.
    \item We propose a teacher-student architecture to alleviate distributional misalignments between the source (or data for pre-training) and target domains (or data for finetuning), followed by regularizations to align representations of the teacher and student networks.
    \item We conduct extensive experiments with driving benchmarks, including Waymo Open dataset and KITTI, to demonstrate the effectiveness of our proposed method.
\end{itemize}

\section{RELATED WORK}
\label{sec:related}


\subsection{Self-Supervised Pre-Training in 3D Object Detection}
Self-supervised pre-training~\cite{GPC, calico, patchcontrast} has drawn considerable attention in contemporary research on LiDAR-based 3D object detection, due to its efficacy in learning point cloud representation without labels and transferability to downstream task with small data.
Notably, GCC-3D~\cite{gcc3d} introduces a framework incorporating geometry-aware contrast in contrastive learning paradigm. PointContrast~\cite{xie2020pointcontrast}, ProposalContrast~\cite{proposalcontrast} leverage point-level and region-level contrast to find correlation between different views.  
In the context of masked autoencoders (MAE), MAEs such as Voxel-MAE~\cite{voxelmae}, Occupancy-MAE~\cite{O-MAE} employ voxel-level masking to reconstruct masked points with decoder.
Recently, GD-MAE~\cite{gdmae} and MV-JAR~\cite{MV-JAR} adopt masking approaches based on transformer architectures. 
Different from previous works that pretrain and finetune with the same dataset, AD-PT~\cite{AD-PT} proposes a diversity-based pretraining on ONCE~\cite{ONCE} dataset to learn unified representations, enabling finetuning on multiple AD datasets. 
Though impressive, AD-PT suffers from suboptimal performance during finetuning stage due to ill-posed domain shift between LiDAR datasets.

\subsection{Unsupervised Domain Adaptation in Point Clouds}
To adapt a source trained 3D LiDAR-based detector to unseen target domain, Unsupervised Domain Adaptation (UDA) addresses the domain gap between labeled source domain and unlabeled target domain. 
Wang, et al~\cite{SN} analyzes variance of object sizes between source and target domains and proposes a statistical normalization to handle the gap. 
3D-COCO~\cite{3dcoco} proposes a contrastive co-training using bird's eye view (BEV) features to progressively learn transferable knowledge. ST3D~\cite{st3d} leverages self-training to reduce source domain bias and enhance quality of pseudo labels in target domain. 
However, prior works demonstrate constrained performance since they overlook beam-induced domain gap. 
LiDAR Distillation~\cite{lidar-distill} addresses the discrepancy in LiDAR beams by generating a pseudo low-beam data by downsampling and transferring knowledge of source model from a high-density data to a low-density data. 
DTS~\cite{dts} extends to various settings of point cloud densities including low-to-high density adaptation by proposing Random Beam Random Sampling and object-graph consistency to match the density of the source domain and target domain.


\subsection{General Model Finetuning}
After many pretraining algorithms with large amounts of unlabeled data are proposed, recent literature also focuses on transferring the general pre-trained model's representation to downstream tasks. 
SCL~\cite{SCL}, Bi-tuning~\cite{Bi-tuning}, Core-tuning~\cite{core-tuning}, and COIN~\cite{COIN} propose finetuning methods using supervised contrastive loss to improve performance in classification tasks.
Li et al.~\cite{Li-et-al} presents L2 norm regularize of parameters between pre-trained and downstream model to improve performance, and AT~\cite{AT}, DELTA~\cite{DELTA} presents behavior-based regularization loss that uses attention to reduce feature map discrepancy. 
DR-Tune~\cite{DR-TUNE} selects features with semantics from the pre-trained model's general features using semantic calibration, presents a distribution regularization method using labels, and demonstrates performance improvement. 
However, most of the above works are studied in 2D classification, and a general model finetuning method has not been proposed for 3D object detection tasks. 
We confirm the existence of a density domain shift between the pretrain and finetuning datasets. Thus we propose a general finetuning framework (DADT) with limited data in 3D object detection by bridging domain gaps.


\section{METHODOLOGY}
\label{sec:method}


\begin{figure*}[t] 
        \centering
        \includegraphics[width=0.9\linewidth]{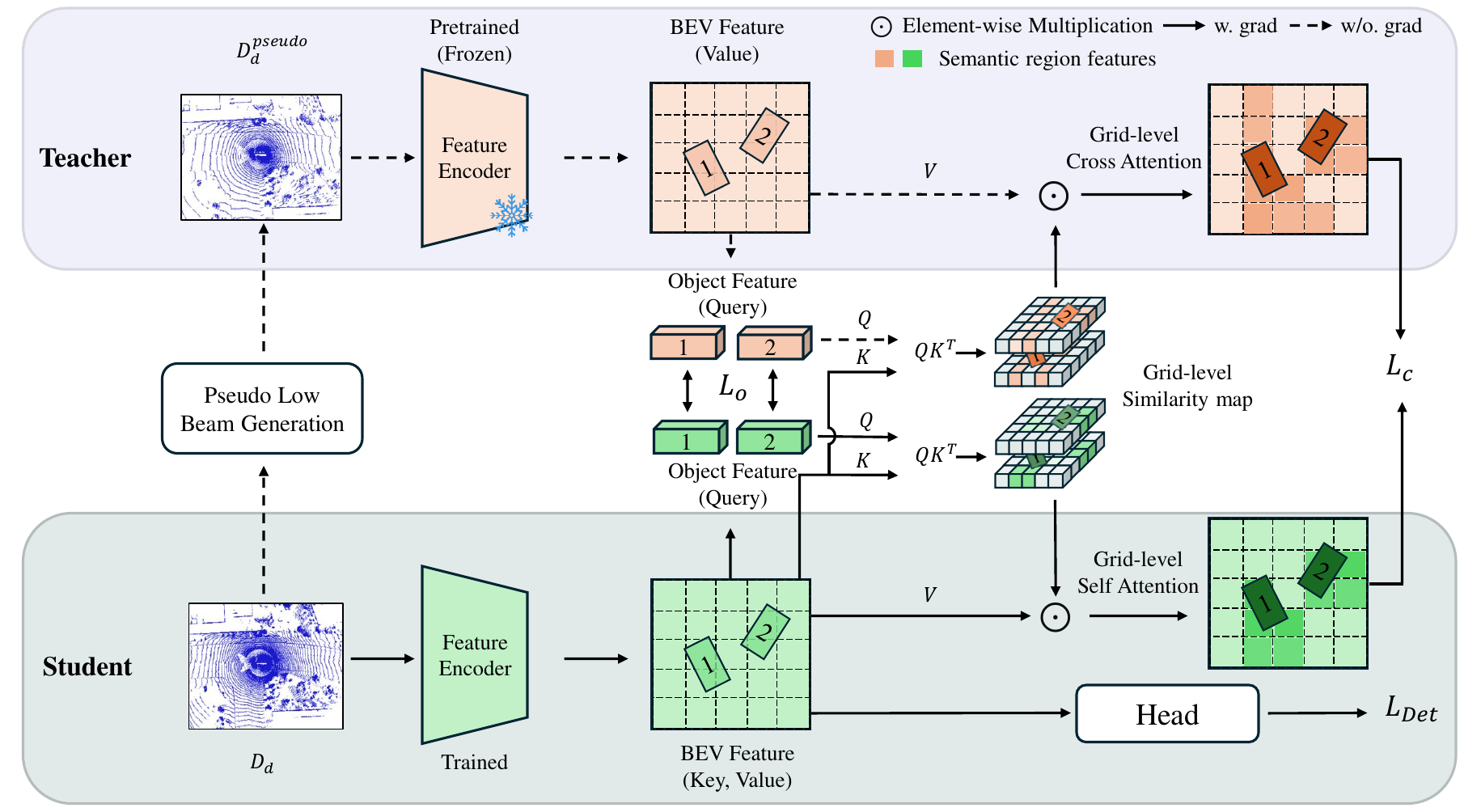}
        \label{fig:main_figure}
    \caption{\textbf{Overview of our proposed Domain Adaptive Distill-Tuning (DADT) framework.} DADT has a teacher-student architecture for reducing a density-driven representational gap. Downstream dataset $D_d$ is downsampled using Pseudo Low Beam Generation to create a $D^{pseudo}_{d}$ similar to pretrain dataset. To supervise and regularize the student's BEV feature distribution to match the teacher's general representation feature distribution, BEV object similarity loss is used to make the same objects in the teacher and student have similar features, and BEV context similarity loss is used to find the grid-similarity of the objects to highlight the semantic features of the objects in the current scene and make them similar.}
    \vspace{-1.0em}
\end{figure*}

\subsection{Problem Statement}

The goal is to solve the domain shift with a few downstream data and proceed with the downstream task under the assumption that there is a beam-induced discrepancy between the pretrain domain and the downstream task domain. Let the pretrain dataset be $D_p$ and the downstream dataset be $D_d = \{(X_i, Y_i)\}_{i=1}^N$ ($N$ is small). 
$N$ is the sample number of the downstream domain $D_d$ while $X_i$ and $Y_i$ denote $i\,$th point cloud and its 3D bounding box label (center, dimensions, heading), respectively. 
3D object detection model consists of 3D encoder, BEV encoder, and detection head. We define a pre-trained model's backbone as $f_{{\theta}^p}(\cdot)$, consisting of 3D and BEV encoders, and its detection head as $g_{{\phi}^p}(\cdot)$, which are parameterized by $\theta^p$ and $\phi^p$ respectively. Then, we aim to finetune a downstream model $f_{{\theta}^d} \cdot g_{{\phi}^d}(\cdot)$, where $f_{{\theta}^d}$ is initialized by $f_{{\theta}^p}$, and $g_{{\phi}^d}$ is randomly initialized.

\myparagraph{Vanilla Finetuning.}
Vanilla finetuning initializes $f_{{\theta}^d} $ as $f_{{\theta}^p}$, randomizes $g_{{\phi}^d}$, and proceeds learning directly. Vanilla optimizes the following objective function.

\begin{equation}
    \boldsymbol{\theta}_*^d, \:	\boldsymbol{\phi}_*^d=\arg \min _{\boldsymbol{\theta}^d, \boldsymbol{\phi}^d} \mathrm{L}\left(f_{\boldsymbol{\theta}^d} \cdot g_{\boldsymbol{\phi}^d} ; \boldsymbol{D_d}\right)
\end{equation}

However, due to the density domain shift, $f_{{\theta}^p}$ cannot learn to represent the BEV feature $F$ for $D_d$ well in limited data. Therefore, it is necessary to supervise $f_{{\theta}^d} $ to represent $D_d$ feature well while utilizing the representation of $f_{{\theta}^p}$ as an encoder. 

\begin{figure}[t] 
        \centering
        \includegraphics[width=0.95\linewidth]{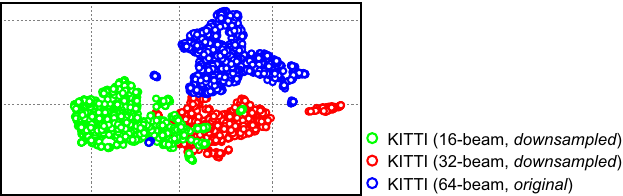}
    \caption{\textbf{UMAP~\cite{umap} Visualizations.} We compare representational differences depending on the beam density (i.e., 16, 32, 64-beams). We use KITTI data, downsampling it into a lower-density LIDAR point cloud.}
    \label{fig:beam}
    \vspace{-1.5em}
\end{figure}

\subsection{Teacher-Student Architecture for Reducing a Density-driven Representational Gap}
Recent studies have repeatedly reported a potential domain shift by differences in the point density of a LiDAR sensor. Our experiment also confirms this as shown in \cref{fig:beam}, where we visualize embeddings from a pre-trained backbone by UMAP~\cite{umap} with different point-density inputs. This necessitates aligning the point density between data for pre-training and finetuning. To address this issue, we utilize the teacher-student learning strategy, where a teacher network takes the density-aligned LiDAR inputs, and a student network uses the original density-non-aligned LiDAR inputs. Formally, we downsample $D_d$ by adopting a pseudo low beam data generation approach~\cite{lidar-distill} to create $D_d^{pseudo}$, thus having a point density similar to $D_p$.



Specifically, we first transform LiDAR points from Cartesian to Spherical coordinates system:
\begin{equation}
    r=\sqrt{x^2+y^2+z^2}, ~
    \phi =\arctan \frac{z}{\sqrt{x^2+y^2}},~ 
    \theta =\arcsin \frac{y}{\sqrt{x^2+y^2}}
    \label{eq:polar}
\end{equation}
where $x, y, z$ are Cartesian coordinates of $D_d$, and $\phi$, $\theta$ are inclination and azimuth angles, respectively. Given $\phi$, we use K-means clustering to assign points to beam labels and uniformly downsample beam clusters along $\phi$. Such downsampled points are then reverted into Cartesian coordinates, yielding $D_d^{pseudo}$. 


\subsection{BEV-based Similarity Losses}
Given the teacher-student networks, the teacher network utilizes the frozen pre-trained backbones with density-aligned inputs, and the student finetunes its backbone (initialized from the pre-trained model) with density-non-aligned inputs. We use the following two regularization losses, i.e., (1) object similarity loss and (2) context similarity loss, to finetune the student network, retaining its representational distribution similar to the teacher network and preventing from overfitting.

\myparagraph{Object Similarity Loss.}
Inspired by recent LiDAR-based studies~\cite{3dcoco, lidar-distill, dts}, we apply the following object similarity loss:
\begin{equation}
    \mathrm{L}_{o}= \sum_{c\in\mathrm{C}}\frac{1}{N_c}\sum_{i=1}^{N_c}\left\|z^{\mathrm{T}}_i - z^{\mathrm{S}}_i\right\|_2
    \label{eq:gt loss},
\end{equation}
where $N_c$ is the number of objects for a given class $c\in\mathrm{C}$. The object BEV features $z^{T}_i$ and $z^{S}_i$ for an object $i$ are extracted from feature encoders of the teacher $\mathrm{T}$ and student $\mathrm{S}$ networks, respectively. We use a ground truth bounding box to extract the features of each agent from BEV representation. Note that we use the sum of per-class normalized losses to address a class imbalance problem during the finetuning step.

\myparagraph{Context Similarity Loss.}
Further, in addition to object-wise similarity loss, we use attention-based context similarity loss $\mathrm{L}_c$ as follows:
\begin{equation}
    \mathrm{L}_{c} =\sum_{c\in\mathrm{C}} \texttt{MSE}(a^\mathrm{T}_c, a^\mathrm{S}_c)
    \label{eq:Attention_loss}
\end{equation}
where $a^\mathrm{T}_c$ and $a^\mathrm{S}_c$ are the average of the attended BEV features for a given class $c\in\mathrm{C}$ from the teacher and student networks, respectively. Formally, we use an attention~\cite{vaswani2017attention} module to obtain the attended feature $a^\mathrm{S}_c$ from the student network: i.e., the object BEV feature $z^{\mathrm{S}}_i$ is used as the query vector, and the student's BEV features of each grid $F^\mathrm{S}\in\mathrm{R}^{h\times w\times d}$ are used as the key and the value vectors as follows: 
%

\begin{equation}
     a^\mathrm{S}_c = F^{\mathrm{S}} \odot\left(\frac{1}{N_c} \sum_{i=1}^{N_c}z^{\mathrm{S}}_i \cdot F^{\mathrm{S}}\right)
    \label{eq: Attention Student Feature}
\end{equation}

%
Similarly, we compute $a^\mathrm{\mathrm{T}}_c$ by applying cross-attention with the following query, key, and value vectors.
%
%

\begin{equation}
     a^\mathrm{T}_c = F^{\mathrm{T}} \odot\left(\frac{1}{N_c} \sum_{i=1}^{N_c}z^{\mathrm{T}}_i \cdot F^{\mathrm{S}}\right)
    \label{eq: Attention Teacher Feature}
\end{equation}

Specifically, we can get the grid-level similarity map for an object by performing a dot product between the object BEV feature and the whole BEV feature. Then, we can calculate the grid-level similarities for all objects in class $\mathrm{C}$ and average these similarities, called Context Similarity. With element-wise multiplication of the BEV feature and Context similarity for class $\mathrm{C}$, we can get the features that emphasize the semantic region by multiplying the similarity of the values. Therefore, we can regularize the student network to focus on the semantic region features and retain its representational distribution similar to the teacher network by applying the MSE loss of $a^\mathrm{T}_i$ and $a^\mathrm{S}_i$.

\myparagraph{Loss Function.}
Ultimately, we minimize the following loss function $\mathrm{L}$: 
\begin{equation}
    \mathrm{L} = \mathrm{L}_{Det} + \lambda_{c} \mathrm{L}_{c} + \lambda_{o} \mathrm{L}_{o}
    \label{eq:loss_sum}
\end{equation}
where $\lambda_{c}$ and $\lambda_{o}$ are hyperparameters to control the weight of each term. 


\section{EXPERIMENTS}
\label{sec:exp}



%
\begin{figure*}[t]
 \begin{minipage}[h]{.55\linewidth}
    \setlength{\tabcolsep}{4pt}
    \renewcommand{\arraystretch}{1} 
    \centering
        \captionof{table}{\textbf{3D Detection Accuracy Comparison with Limited Data.} We compare the detection performance of finetuned models with limited data, from 32 to 128, in terms of two metrics: AP and APH. Note that we use the Waymo~\cite{waymo} validation set to measure scores. All models are based on the SECOND~\cite{second}.}
         \label{tab:waymo}
    \resizebox{\linewidth}{!}{%
    \begin{tabular}{lccccc}
    \toprule
        \multirow{2}{*}{Finetuning} &
        \multirow{2}{*}{\# of Data} & 
        \multicolumn{4}{c}{AP$\uparrow$ / APH$\uparrow$} \\
        \cmidrule{3-6}
        & & Overall & Vehicle & Pedestrian & Cyclist \\\midrule
        Baseline & 32 & 00.00 / 00.00 & 00.00 / 00.00 & 00.00 / 00.00& 00.00 / 00.00 \\     
        \rowcolor{LightGrey} DADT (ours) & 32 & 04.44 / 03.81 & 10.91 / 10.66 & 01.66 / 01.15 & 00.77 / 00.71 \\
        \rowcolor{LightGrey} &  & (\red{4.44$\uparrow$ / 3.81$\uparrow$})  & (\red{10.91$\uparrow$ / 10.66$\uparrow$}) & (\red{1.66$\uparrow$ / 1.15$\uparrow$}) & (\red{0.77$\uparrow$ / 0.71$\uparrow$})\\
        \midrule  
        Baseline & 64 & 02.75 / 02.71 & 05.98 / 05.90 & 00.01 / 00.00 & 02.26 / 02.23 \\ 
        \rowcolor{LightGrey} DADT (ours) & 64 & 23.73 / 21.48 & 25.74 / 25.29 & 17.05 / 11.64 & 28.39 / 27.51  \\
        \rowcolor{LightGrey} &  & (\red{20.98$\uparrow$ / 18.77$\uparrow$})  & (\red{19.76$\uparrow$ / 19.39$\uparrow$}) & (\red{17.01$\uparrow$ / 11.64$\uparrow$}) & (\red{26.13$\uparrow$ / 25.28$\uparrow$})\\\midrule 
        Baseline & 96 & 15.02 / 14.11 & 14.76 / 14.51 & 07.22 / 05.21& 23.07 / 22.62 \\ 
        \rowcolor{LightGrey} DADT (ours) & 96 & 33.77 / 30.55 & 37.65 / 37.03 & 26.33 / 18.30 & 37.34 / 36.32 \\
        \rowcolor{LightGrey} &  & (\red{18.75$\uparrow$ / 16.44$\uparrow$})  & (\red{22.89$\uparrow$ / 22.52$\uparrow$}) & (\red{19.11$\uparrow$ / 13.09$\uparrow$}) & (\red{14.27$\uparrow$ / 13.7$\uparrow$})\\\midrule  
        Baseline & 128 & 33.07 / 30.34 & 34.96 / 34.35 & 27.26 / 20.52 & 37.00 / 36.16 \\ 
        \rowcolor{LightGrey} DADT (ours) & 128 & 40.22 / 36.75 & 46.48 / 45.71 & 33.33 / 24.67 & 40.85 / 39.89 \\\rowcolor{LightGrey} &  & (\red{7.15$\uparrow$ / 6.41$\uparrow$})  & (\red{11.52$\uparrow$ / 11.36$\uparrow$}) & (\red{6.07$\uparrow$ / 4.15$\uparrow$}) & (\red{3.85$\uparrow$ / 3.73$\uparrow$})\\
        \bottomrule 
        \end{tabular}}
  \end{minipage}
\hspace{1em}
\begin{minipage}[h]{.4\linewidth}
    \setlength{\tabcolsep}{6pt}
    \renewcommand{\arraystretch}{1} 
    \centering
    \captionof{table}{\textbf{3D Detection Accuracy Comparison on KITTI Dataset.} We report scores in terms of 3D Average Precision (AP), with the IoU threshold set to 0.7 for Cars and 0.5 for Pedestrians and Cyclists.}\label{tab:kitti}
    \resizebox{\linewidth}{!}{%
    \begin{tabular}{lccc}
    \toprule
        \multirow{2}{*}{Finetuning} & \multirow{2}{*}{\# of Data} & \multicolumn{2}{c}{mAP (Mod.)$\uparrow$}\\ \cmidrule{3-4}
        && SECOND & PV-RCNN++\\\midrule
        Baseline & 1\% (37) & 16.76 & 15.85\\    
        \rowcolor{LightGrey} DADT (ours) & 1\% (37) & 21.66 & 20.21\\ 
        \rowcolor{LightGrey} & & (\red{4.90$\uparrow$}) & (\red{4.36$\uparrow$})\\ \midrule
        Baseline & 2\% (74)& 29.60 & 40.94 \\  
        \rowcolor{LightGrey}  DADT (ours) & 2\% (74) & 31.28 & 44.49 \\
        \rowcolor{LightGrey} & & (\red{1.68$\uparrow$}) & (\red{3.55$\uparrow$})\\ \midrule
        Baseline & 3\% (111)& 38.91 & 49.53 \\ 
        \rowcolor{LightGrey}  DADT (ours) & 3\% (111)& 42.24 & 54.73\\ 
        \rowcolor{LightGrey} & & (\red{3.33$\uparrow$}) & (\red{5.20$\uparrow$})\\
    \bottomrule 
    \end{tabular}}
  \end{minipage}
  \vspace{-1.5em}
\end{figure*}


\myparagraph{Datasets.}
To evaluate the effectiveness of our proposed method, we use two widely-used public datasets: KITTI~\cite{kitti} and Waymo Open Dataset~\cite{waymo}. The former was collected with a 64-beam Velodyne LiDAR sensor in Germany, providing  7,481 annotated LiDAR frames (3,712 for training and 3,769 for validation). The latter contains annotated 19M LiDAR frames (15M for training and 4M for testing) collected with multiple sensors (a single 64-beam and four 200-beam LiDAR sensors). A subset of a few frames are uniformly sampled to evaluate our model under limited data scenarios. 

\myparagraph{Baseline Models.}
While our method generally applies to various 3D LiDAR-based object detection models without notable restrictions, we use the following two commonly-used detectors: SECOND~\cite{second}, PV-RCNN++~\cite{pvrcnn++}. Also, we utilize the pre-trained AD-PT~\cite{AD-PT} model trained on ONCE~\cite{ONCE} dataset as our Oracle model. Note that ONCE dataset is a large-scale driving dataset collected in China with a 40-beam LiDAR, including various environmental conditions (e.g., day/night and sunny/rainy scenes). This model is ideal for our evaluation since (i) pre-trained models are publicly available for researchers to easily access for reproduction and (ii) ONCE dataset has a potential domain gap with our evaluation datasets (i.e., KITTI and Waymo Open Dataset) due to their differences in sensor configurations (40-beam LiDAR vs. 64-beam and 200-beam customized LiDAR sensors) and locations (China vs. Germany and USA). we finetune these pre-trained models with limited amounts of target data, denoting them as baselines.

\myparagraph{Implementation Details.}
To finetune LiDAR-based 3D object detectors, we set $\lambda_{atten} = 1.0$ and $\lambda_{gt} = 1.0$, optimized by a grid search. As we randomly select only a few frames to finetune models, the model's performance might be varied regarding randomness. Thus, we report average scores from multiple independent runs for all experiments. We conduct each experiment on four NVIDIA 3090 GPUs with a batch size of 32 and 12 for KITTI and Waymo, respectively. Following the recent work, i.e., 3DTrans~\cite{3dtrans2023}, we use the same augmentation and optimization techniques for all models. 


\begin{figure}[t] 
        \centering
        \includegraphics[width=\linewidth]{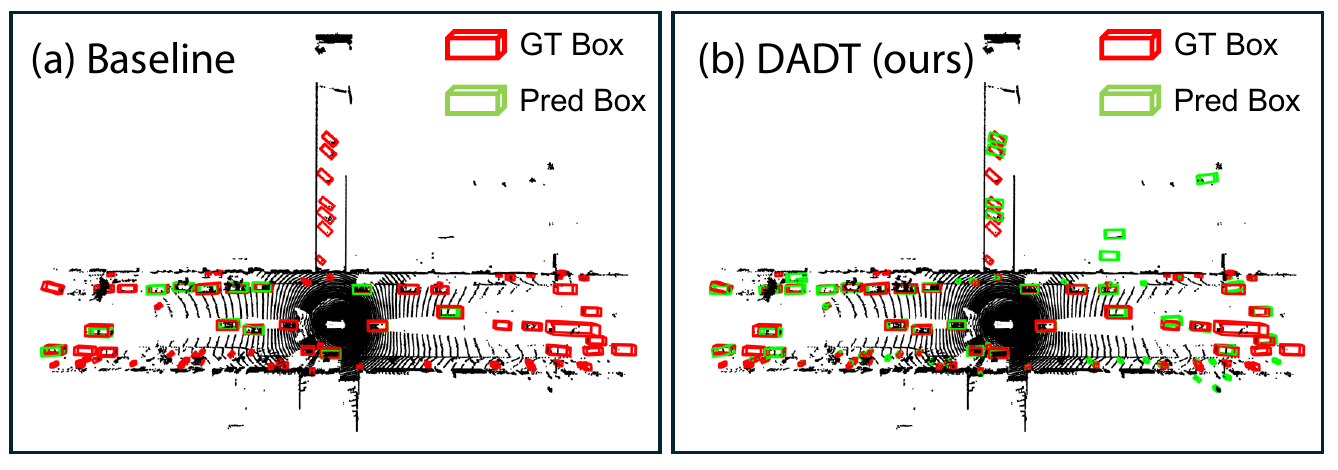}
    \caption{\textbf{Examples of Detected 3D Objects from (a) baseline model and (b) ours.} Red boxes and green boxes denote ground-truth and predicted bounding boxes, respectively.}
    \label{fig:qualitative}
    \vspace{-1.5em}
\end{figure}

\myparagraph{Evaluation Metrics.}
We use the standard practice in evaluating LiDAR-based 3D object detectors regarding datasets. With KITTI datasets, we report the mean Average Precision (mAP) using 40 recall positions for three categories (i.e., cars, pedestrians, and cyclists) for easy, moderate, and hard cases. For the Waymo Open Dataset, we use Average Precision (AP) and Average Precision with Heading (APH) under Level 1 setting for vehicles, pedestrians, and cyclists.


\subsection{Effect of Finetuning Pre-trained Model with Limited Data}

\begin{figure}[t] 
        \centering
        \includegraphics[width=0.9\linewidth]{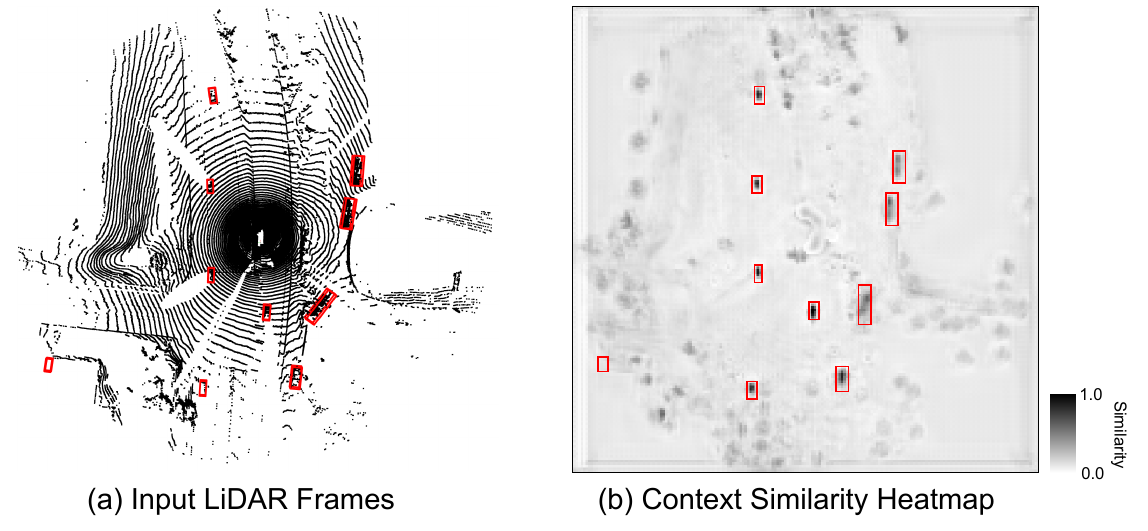}
    \caption{\textbf{Example of Context Similarity Heatmap.} Red boxes denote ground-truth bounding boxes.}
    \label{fig:attention-viz}
    \vspace{-1.5em}
\end{figure}

\myparagraph{Evaluation on Waymo Open Dataset.}
We start to evaluate the effectiveness of our approach by measuring detection accuracies on the Waymo~\cite{waymo} validation set regarding two primary metrics (i.e., AP and APH), as reported in \cref{tab:waymo}. Based on our baseline model, which is built upon SECOND architecture pre-trained with the AD-PT method on the ONCE~\cite{ONCE} dataset, we compare our proposed finetuning approach with conventional common practice, denoted as a vanilla model. We report 3D detection scores with different numbers of limited data, i.e., from 32 to 128 LiDAR frames ra ndomly sampled from training dataset, evaluating with the 20\% validation set. We observe in \cref{tab:waymo} that our method consistently shows improvements by a large gap, 4.44--20.98 and 3.81--18.77 higher AP and APH, respectively. This confirms that our method is effective for finetuning a pre-trained model with limited data. In \cref{fig:qualitative}, we also compare the 3D box prediction results on Waymo validation set with Vanilla (baseline) and DADT models trained on 64 frames. As illustrated in Figure \ref{fig:qualitative}, we can see that baseline is able to detect some nearby cars, but not the rest of the objects. However, in case of DADT, it detects not only nearby objects but also distant objects. To demonstrate the effectiveness of our proposed context similarity loss, we also visualize context similarity of the models trained with 96 frames in \cref{fig:attention-viz}. We observe that our DADT highlights objects and its nearby environment.


\begin{figure}[t] 
        \centering
        \includegraphics[width=0.9\linewidth]{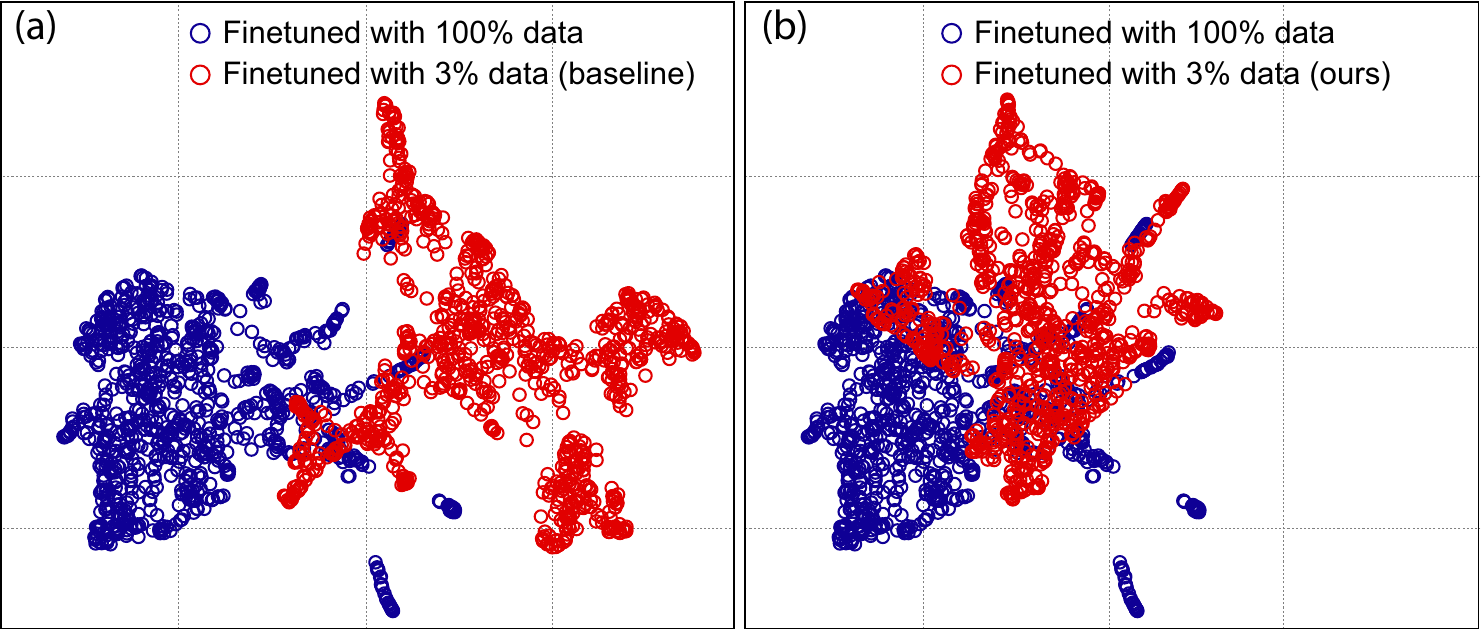}
    \caption{\textbf{UMAP Visualizations} of the Oracle (a pre-trained backbone finetuned with 100\% KITTI training data) and those from our baseline and ours. KITTI validation set is used for visualization.}
    \label{fig:compare}
    \vspace{-1.5em}
\end{figure}

\myparagraph{Evaluation on KITTI Dataset.}
Further, our experiment with the KITTI~\cite{kitti} dataset shows significant gains consistent with our Waymo Open dataset results. As shown in \cref{tab:kitti}, we evaluate models with different numbers of LiDAR frames (from a training set) and tested with the whole validation set. In this experiment, we use two kinds of 3D LiDAR-based object detection, i.e., SECOND~\cite{second} and PV-RCNN++~\cite{pvrcnn++}, which are pre-trained with AD-PT on ONCE~\cite{ONCE} dataset.


\myparagraph{UMAP Analysis.}
Our proposed model regularizes the finetuning process by constraining the representations of the LiDAR point cloud so that they are similar to those of pre-trained models. To verify this, we use UMAP~\cite{umap} to visualize and compare embeddings. As expected, we observe in \cref{fig:compare} that a naive supervised finetuning often guides the model to produce distributional shifts, overfitting to the small dataset (thus, will not generalize well to unseen data). However, our approach guides the model to produce representational distributions similar to the Oracle model (supervised finetuned model with 100\% LiDAR frames), preventing overfitting.

{
\setlength{\tabcolsep}{3pt}
\renewcommand{\arraystretch}{1} 
\begin{table}[t]
	\begin{center}
        \caption{\textbf{Ablation Study.} We evaluate the effect of finetuned models with and without the following three components: (i) Pseudo Low Beam Generation (PLBG), (ii) the use of object similarity loss $\mathrm{L}_{o}$ and (iii) context similarity loss $\mathrm{L}_{c}$. Data: Waymo~\cite{waymo}. PLBG$^\dagger$: Pseudo Low Beam Generation.}
        \label{tab:ablation}
    	\resizebox{\linewidth}{!}{%
    	\begin{tabular} {@{}ccc|cccc@{}} \toprule
            \multirow{2}{*}{PLBG$^\dagger$} & \multicolumn{2}{c|}{Regularizer} & \multicolumn{4}{c}{AP$\uparrow$} \\\cmidrule{2-7}
            & Use of $\mathrm{L}_{o}$ & Use of $\mathrm{L}_{c}$ & Vehicle & Pedestrian & Cyclist & Avg. \\ \midrule
            \rowcolor{LightGrey} \cmark & \cmark & \cmark & 37.65 & 26.33 & 37.34 & 33.77 \\\midrule
            \cmark & \cmark & \xmark &35.69 & 24.72 & 35.61 & 32.01 (1.76$\downarrow$)\\
            \cmark & \xmark & \cmark   &36.26 &24.52 &35.78 & 32.19 (1.58$\downarrow$)\\
            \xmark & \cmark & \cmark  & 36.55 & 26.19 &36.43 & 33.06 (0.71$\downarrow$)\\
            \midrule
            \xmark & \xmark & \xmark &14.76 & 7.22 & 23.07 & 15.02 (18.75$\downarrow$)\\
            \bottomrule
            \end{tabular}}
        
     \end{center}\vspace{-1.5em}
\end{table}
}
%
%
\myparagraph{Effect of Each Component.} 
In \cref{tab:ablation}, we provide our ablation study results, which show the effect of our module and two regularization functions: (i) Pseudo Low Beam Generation, (ii) the use of $\mathrm{L}_{o}$ and (iii) $\mathrm{L}_{c}$. We use the Waymo Open dataset (finetuning with 96 LiDAR frames from the training set and evaluating with the 20\% validation set). Our experiment shows that each building block has meaningful contributions, while their combinations offer significant synergy, outperforming the baseline with a large gap (compare 2-4th rows vs. bottom row).    


{
\setlength{\tabcolsep}{8pt}
\renewcommand{\arraystretch}{1} 
\begin{table}[t]
    \centering
    \caption{\textbf{Performance of Continual Finetuning.} Given a model finetuned on the Waymo Open dataset (with 96 LiDAR frames from the training set), we further evaluate the ability to adapt to a new dataset (i.e., KITTI) continuously. Note that direct transfer indicates a model that is first trained on the Waymo Open dataset, directly followed by training on the KITTI dataset.}
    \label{tab:DA}
    \resizebox{\linewidth}{!}{%
    \begin{tabular} {@{}lccc@{}} \toprule
        \multirow{2}{*}{Waymo $\rightarrow$ KITTI }  & \multicolumn{3}{c}{Car AP$\uparrow$} \\ \cmidrule{2-4}
          & Easy & Moderate & Hard \\ \midrule
        Direct Transfer  &78.05 & 63.67 & 59.35\\
        Baseline  &81.19 & 64.02 & 59.54 \\
        \rowcolor{LightGrey} DADT (ours) & 83.84 (\red{2.65$\uparrow$})  & 66.07 (\red{2.05$\uparrow$}) &60.89 (\red{1.35$\uparrow$}) \\
        \bottomrule
    \end{tabular}}
    \vspace{-0.5em}
\end{table}
}

{
\setlength{\tabcolsep}{5pt}
\renewcommand{\arraystretch}{1} 
\begin{table}[t]
	\begin{center}
        \caption{\textbf{KITTI Performance Comparison in Semi-supervised Learning Setting.} All models are based on PV-RCNN++~\cite{pvrcnn++} architecture and initialized from the pre-trained model with AD-PT~\cite{AD-PT} on the ONCE~\cite{ONCE} dataset, except the scratch model, which is trained from random initialization. \textit{Abbr.} S: Supervised (use the whole 3,712 training data), SS: Semi-supervised (use the combination of the few 111 labeled data and 3,712 pseudo-labeled data).}
        \label{tab:semi-sup}
    	\resizebox{0.95\linewidth}{!}{%
    	\begin{tabular} {@{}lccccc@{}} \toprule
            \multirow{2}{*}{Model} & \multirow{2}{*}{S$|$SS} & \multicolumn{4}{c}{mAP (Mod.)$\uparrow$} \\ \cmidrule{3-6}
            & & Car & Pedestrian & Cyclist & Avg. \\ \midrule
            Baseline & SS & 76.68 & 53.71 & 68.10 & 66.83\\
            \rowcolor{LightGrey} DADT (ours) & SS & 79.47  & 55.67 & 69.72 & 68.29\\
            \rowcolor{LightGrey} &  & (\red{2.79$\uparrow$}) & (\red{1.96$\uparrow$})& (\red{1.62$\uparrow$})& (\red{1.46$\uparrow$})\\\midrule
            Scratch & S  & 84.51 & 56.93 & 71.09 & 70.84\\\bottomrule
            \end{tabular}}
     \end{center}\vspace{-2em}
\end{table}
}

\subsection{Performance in Continual Training Scenarios}

\myparagraph{Continual Finetuning.} 
Further, as our proposed approach keeps the original representational distribution while learning new data, our model might be advantageous for continuously adapting various datasets sequentially, learning more robust yet effective domain-invariant features. To evaluate the model's ability to adapt continually, we experiment with a scenario where a model is first trained on the Waymo Open dataset (with 96 LiDAR frames) with EMA~\cite{ema}, followed by training on the KITTI dataset (with 111 LiDAR frames). As shown in \cref{tab:DA}, our model provides better detection performance than direct transfer and vanilla supervised finetuned models. 


\myparagraph{Semi-supervised Learning.}
We also evaluate the performance in the semi-supervised learning setup, where we first pretrain the model using 3\% labeled data from KITTI~\cite{kitti} training set and conduct further training by creating pseudo labels based on ST3D~\cite{st3d} for the rest of training data. As we summarized in \cref{tab:semi-sup}, our model shows better performance than our baseline model, showing closer to a model, which is trained from scratch with 100\% labeled data (68.29 vs. 70.84 in terms of avg. mAP. Compare 2nd and 3rd rows). This demonstrates that it is possible to maintain high performance by labeling only limited data and then learning through self-training on unlabeled data. It also suggests that in the real industry, high-performance models can be made by online learning by finetuning a model with limited labels and self-training with the online LiDAR frames collected.




\section{CONCLUSION}
In this paper, we introduce DADT, a distillation based domain adaptive finetuning framework for 3D LiDAR-based detection under limited target data. Our framework employs pseudo beam generation and novel BEV attention-based regularizer to effectively alleviate serious domain shift present in the finetuning of general pre-trained detection model with limited data. We comprehensively validate the effectiveness and practicality of our framework as it substantially improves performance of various baselines on Waymo and KITTI datasets and can be applicable to different problem settings.










\small{
\section*{ACKNOWLEDGMENT}
This work was supported by Autonomous Driving Center, Hyundai Motor Company R\&D Division. This work was supported by Basic Science Research Program through the National Research Foundation of Korea funded by the Ministry of Education(NRF-2021R1A6A1A13044830, 15\%) and supported by Institute of Information \& communications Technology Planning \& Evaluation grant funded by the Korea government(MSIT) (RS-2022-II220043, 15\%, IITP-2024-RS-2024-00397085, 15\%).
}


\bibliographystyle{IEEEtran}
\bibliography{IEEEabrv,ref}

\end{document}